# Stochastic Optimization Algorithms

Pierre Collet
Université du Littoral Côte d'Opale
Laboratoire d'Informatique du Littoral BP 719
62100 Calais cedex — France
pierre.collet@Univ-Littoral.fr

Jean-Philippe Rennard*
Grenoble Graduate School of Business
12, rue Pierre Sémard BP 127
38003 Grenoble cedex 01 — France
jp@rennard.org

**Abstract**: When looking for a solution, deterministic methods have the enormous advantage that they do find global optima. Unfortunately, they are very CPU-intensive, and are useless on untractable NP-hard problems that would require thousands of years for cutting-edge computers to explore.
In order to get a result, one needs to revert to stochastic algorithms, that sample the search space without exploring it thoroughly. Such algorithms can find very good results, without any guarantee that the global optimum has been reached; but there is often no other choice than using them.
This chapter is a short introduction to the main methods used in stochastic optimization.

## Introduction

The never ending search for productivity has made optimization a core concern for engineers. Quick process, low energy consumption, short and economical supply chains are now key success factors

Given a space $\Omega$ of individual solutions $\omega \in R^n$ and an objective function $f$, $f(\omega) \to R$, optimizing is the process of finding the solution $\omega^*$ which minimizes (maximizes) $f$.

For hard problems, optimization is often described as a walk in a *fitness landscape*. First proposed by biologist S. Wright in 1932 (Wright, 1932), fitness landscapes aimed at representing the fitness of a living organism according to the genotype space. While optimizing, *fitness* measures the quality of a solution, and fitness landscapes plot solutions and corresponding goodness (fitness). If one wishes to optimize the function $x+1=0$, then depending on the choice of the error measure, fitness can for example be defined as $-|-(x+1)|$ or as $1/|-(x+1)|$. The optimization process then tries to find the peak of the fitness landscape (see figure 1-a).

<<INSERT FIGURE 1>>

This example is trivial and the optimum is easy to find. Real problems are often multimodal, meaning that their fitness landscapes contain several local optima (*i.e.* points whose all neighbors have a lower fitness see figure 1-b). This is particularly true when variables interact with one another (*epistatis*).

Usual analytical methods, like gradient descent, are often unable to find a global optimum, since they are unable to deal with such functions. Moreover, companies mostly deal with combinatorial problems like quadratic assignment, timetabling or scheduling problems. These problems using discrete states generate non-continuous objective functions that are unreachable through analytical methods.

Stochastic optimization algorithms were designed to deal with highly complex optimization problems. This chapter will first introduce the notion of complexity and then present the main stochastic optimization algorithms.

# NP-complete problems and combinatorial explosion

In December, *Santa Claus* must prepare the millions of presents he has to distribute for Christmas. Since the capacity of his sleigh is finite, and he prefers to minimize the number of runs, he would like to find the best way to organize the packs. Despite the apparent triviality of the task, *Santa Claus* is facing a very hard problem. Its simplest formulation is the *one-dimensional bin packing problem*. Given a list $L = (a_1, a_2, ..., a_n)$ of *items* with sizes $0 < s(a_i) \leq 1$, what is the minimum number *m* of unit-capacity bins $B_j$ such that $\sum_{a_i \in B_j} s(a_i) \leq 1, 1 \leq j \leq m$ ? This problem is known to be NP-hard (Coffman, Garey, & Johnson, 1996).

Various forms of *bin packing problem* are very common. Transportation industry must optimize truck packing given weight limits, press has to organize advertisements minimizing the space, metal sheet industry must solve the *cutting-stock problem* (how to minimize waste when cutting a metal sheet)…

Such problems are very tough because we do not know how to build algorithms that can solve them in *polynomial-time*; they are said to be *intractable problems*. The only algorithms we know for them need an *exponential-time*. Table 1 illustrates the evolution of time algorithms for polynomial-time problems *vs* non-polynomial. Improving the speed of computers or algorithms is not the solution, since if the speed is multiplied, the gain of time is only additive for exponential functions (Papadimitriou & Steiglitz, 1982).

<<INSERT TABLE 1>>

The consequences of the computational complexity for a great many real world problems are fundamental. Exact method for scheduling problems "become computationally impracticable for problems of realistic size, either because the model grows too large, or because the solution procedures are too lengthy, or both, and heuristics provide the only viable scheduling techniques for large projects" (Cooper, 1976).

*Heuristics and meta-heuristics*

Since many real-world combinatorial problems are NP-hard, it is not possible to guarantee the discovery of the optimum. Instead of exact methods, one usually uses *heuristics*, which are approximate methods using iterative trial and error processes, to approach the best solution. Many of them are nature-inspired, and their latest development is to use *metaheuristics*. "A metaheuristic is an iterative master process that guides and modifies the operations of subordinate heuristics to efficiently produce high-quality solutions. It may manipulate a complete (or incomplete) single solution or a collection of solutions at each iteration. The subordinate heuristics may be high (or low) level procedures, or a simple local search, or just a construction method." (Voss, Martello, Osman, & Roucairol, 1999).
Metaheuristics are high level methods guiding classical heuristics. They deal with a dynamic balance between *diversification* (exploration of the solution space) and *intensification* (exploitation of the accumulated knowledge) (Blum & Roli, 2003).

# Stochastic algorithms

### Random search

Random search is what it says it is. In essence, it simply consists in picking up random potential solutions and evaluating them. The best solution over a number of samples is the result of random search.

Many people do not realize that a stochastic algorithm is nothing else than a random search, with hints by a chosen heuristics (or meta-heuristics) to guide the next potential solution to evaluate. People who realize this feel uneasy about stochastic algorithms, because there is not guarantee that such an algorithm (based on random choices) will *always* find the global optimum.

The only answer to this problem is a probabilistic one:
- If, for a particular problem, one already knows the best solution for different instances of this problem, and
- if, over a significative number of runs, the proposed stochastic algorithm finds a solution that in average is 99% as good as the known optimum for the tested instances of the problem, then,
- one can hope that on a new instance of the problem for which the solution is not known, the solution found by the stochastic algorithm will be 99% as good as the unknown optimum over a significative number of runs.

This claim is not very strong, but there are not many other options available: if one absolutely wants to get *the* global optimum for a large NP-Hard problem, the only way is to let the computer run for several hundred years (cf. table 1)… The stochastic way is therefore a pragmatic one.

### Computational Effort

As can be seen above, it is difficult to evaluate the performance of stochastic algorithms, because, as Koza explains for genetic programming in (Koza, 1994):

> *Since genetic programming is a probabilistic algorithm, not all runs are successful at yielding a solution to the problem by generation $G$.*
> *When a particular run of genetic programming is not successful after the prespecified number of generations $G$, there is no way to know whether or when the run would ever be successful. When a successful outcome cannot be guaranteed for every run, there is no knowable value for the number of generations that will yield a solution […].*

Koza therefore proposes a metrics to measure what he calls the *computational effort* required to solve a problem, that can be extended to any stochastic algorithm where evaluations consume a significant fraction of the computer resources:

One first calculates $P(n)$, the cumulative probability of success by the number of evaluations

$n$, being the number of runs that succeeded on or before the $n$th evaluation, divided by the number of runs conducted.

The computational effort $I(n,z)$ can then be defined as the number of evaluations that must be computed to produce a satisfactory solution with probability greater than $z$ (where $z$ is usually 99%), using the formula: $n * \left\lceil \frac{\ln(1-z)}{\ln(1-P(n))} \right\rceil$.

## No Free Lunch Theorem

Random search is also important because it serves as a reference on which one can judge stochastic algorithms. A very important theorem is that of the *No Free Lunch* (Wolpert & Macready, 1995). This theorem states that no search algorithm is better than a random search on the space of all possible problems —in other words, if a particular algorithm does better than a random search on a particular type of problem, it will not perform as well on another type of problem, so that all in all, its global performance on the space of all possible problems is equivalent to a random search.

The overall implication is very interesting, as it means that an off the shelf stochastic optimizer cannot be expected to give good results on any kind of problem (no free lunch): a stochastic optimizer is not a black box: to perform well, such algorithms must be expertly tailored for each specific problem.

## Hill-climbing

*Hill-climbing* is the basis of most *local search* methods. It is based on:
- A set of feasible solutions $\Omega = \{\omega; \omega \in R^n\}$.
- An objective function $f(\omega)$ that can measure the quality of a candidate solution.
- A neighborhood function $N(\omega) = \{\omega_n \in \Omega \mid dist(\omega_n, \omega) \leq \delta\}$ able to map any candidate solution to a set of close candidate solutions.

The optimization algorithm has to find a solution $\omega^*, \forall \omega \in \Omega, f(\omega^*) \leq f(\omega)$. The basic *hill-climbing* algorithm is trivial:
1. Build a candidate solution $\omega \in \Omega$.
2. Evaluate $\omega$ by computing $f(\omega)$ and set $\omega^* \leftarrow \omega$.
3. Select a neighbor $\omega_n \in N(\omega)$ and set $\omega \leftarrow \omega_n$.
4. If $f(\omega) \leq f(\omega^*)$ set $\omega^* \leftarrow \omega$.
5. If some stopping criterion is met, exit else go to 3.

For example, if one considers the famous Traveling Salesman Problem (TSP: given a collection of cities, finding the shortest way of visiting them all and returning back to the starting point), which is an NP-hard problem (cf. table 1 for complexity). The candidate solution is a list of cities e.g. *F-D-B-A-E-C* and the objective function is the length of this journey. There are many different ways to build a neighborhood function. *2-opt* (Lin, 1965) is one of the simplest since it just reverses a sequence. Applying *2-opt* could lead to F-*E-A-B-D*-C. The new tour will be selected if it is shorter than the previous one, otherwise one will evaluate another neighbor tour.

More advanced *hill-climbing* methods look for the best neighbor:
1. Build a candidate solution $\omega \in \Omega$.
2. Evaluate $\omega$ by computing $f(\omega)$.
3. For each neighbor $\omega_n \in N(\omega)$, evaluate $f(\omega_n)$.

4. If all $f(\omega_n)$ are $\geq f(\omega)$ (local optimum) then exit.
5. Else select $\omega^*, \forall \omega_n \in N(\omega), f(\omega^*) < f(\omega_n)$ as the current candidate solution and set $\omega \leftarrow \omega^*$.
6. Go to 3.

The main advantage of *hill-climbing* is its simplicity, the core difficulty usually being the design of the neighborhood function. The price for this simplicity is a relative inefficiency. It is trivial to see that *hill-climbing* is easily trapped in local minima. If one starts from point A (see figure 1-b), it will not be able to reach the global optimum, since once on top of the first peak, it will not find any better point and will get stuck there.

Even though many advanced forms of *hill-climbing* have been developed, these methods are limited to smooth and unimodal landscapes. "A question for debate in medieval theology was whether God could create two hills without an intervening valley (…) unfortunately, when optimizing functions, the answer seems to be no" (Anderson & Rosenfeld, 1988, p.551). This is why search rules based on local topography usually cannot reach the highest point.

## *Simulated annealing*

*Simulated annealing* (Kirkpatrick, Gellat, & Vecchi, 1983) is an advanced form of *hill-climbing*. It originates in metallurgy. While annealing a piece of metal, quickly lowering the temperature leads to a defective crystal structure, far from the minimum energy level. Starting from a high temperature, cooling must be progressive when approaching the freezing point in order to obtain a nearly-perfect crystal, which is a crystal close to the minimum energy level. Knowing that the probability for a system to be at the energy level $E_0$ is $p(E_0) = \exp(-E_0/k_B T)/Z(T)$, where $k_B$ is the *Boltzmann constant*, $T$ the temperature and $Z(T)$ a normalizing function, Metropolis et al. proposed in 1955 a simple algorithm to simulate the behavior of a collection of atoms at a given temperature (Metropolis, Rosenbluth, Rosenbluth, Teller, & Teller, 1955). At each iteration, a small random move is applied to an atom and the difference of energy $\Delta E$ is computed. If $\Delta E \leq 0$ the new state is always accepted. If $\Delta E > 0$ the new state is accepted according to a probability $p(\Delta E) = \exp(-\Delta E/k_B T)$.

*Simulated annealing* is based on a series of Metropolis algorithms with a decreasing temperature. It can shortly be described this way:
1. Build a candidate solution $\omega \in \Omega$.
2. Evaluate $\omega$ by computing $f(\omega)$.
3. Select a neighbor candidate solution $\omega_n \in N(\omega)$.
4. If $f(\omega_n) \leq f(\omega)$ then set $\omega \leftarrow \omega_n$ and exit if the evaluation is good enough.
5. Else select $\omega_n$ ($\omega \leftarrow \omega_n$) according to the probability: $p = \exp(-(f(\omega_n) - f(\omega))/T_i)$ where $T_i$ is the current temperature which decreases over time.
6. Go to 3.

Uphill moves (step 5) allow overcoming local minima. One can illustrate the difference between *hill-climbing* and *simulated annealing* with the rolling ball metaphor (see Figure 2). Imagine a ball on a bumpy surface. The ball will roll down and stop at the first point of minimum elevation which usually is a local optimum. By tolerating uphill moves, *simulated annealing* somehow "shakes" the surface pushing the ball beyond the local minimum. At the beginning of the process, the surface is brutally shaken —the temperature is high— allowing a large exploration. The reduction of the temperature progressively decreases the shaking to prevent the ball from leaving the global optimum.
<center>**<<INSERT FIGURE 2>>**</center>

*Simulated annealing* is efficient, but slow. Many improvements have been proposed, like the *rescaled simulated annealing* which limits the transitions in a band of energy centered around a target energy level by using $\Delta E_{ij} = (\sqrt{E_j} - \sqrt{E_t})^2 - (\sqrt{E_i} - \sqrt{E_t})^2$, with typically $E_t = \alpha T^2, \alpha > 0$ (Hérault, 2000). This method "flattens" the error surface at the beginning of the process, minimizing the tendency of the algorithm to jump among local minima.

## *Tabu Search*

"Tabu search may be viewed as 'meta-heuristic' superimposed on another heuristic. The approach undertakes to transcend local optimality by a strategy of forbidding certain moves." ((Glover, 1986), this is the first appearance of the term meta-heuristic). Like *simulated annealing*, it is as an advanced form of *hill-climbing*, based on a set of feasible solutions $\Omega$, an objective function $f(\omega)$ and a neighborhood function $N(\omega)$. *Tabu Search* tries to overcome local minima by allowing the selection of non-improving solutions and by using a procedure which avoids cycling moves. Unlike *simulated annealing*, the probability of selection of a non-improving move is not applied to a given neighbor, but to the set of neighbors. To avoid cycles, *Tabu Search* implements a list $T$ of tabu moves which in the basic form contains the $t$ last moves. The simple *Tabu Search* works as follows (Glover, 1989):

    1. Select a potential solution $\omega \in \Omega$ and let $\omega^* \leftarrow \omega$. Initialize the iteration counter $k = 0$ and let $T = \emptyset$.
    2. If $N(\omega) - T = \emptyset$ go to 4. Otherwise, increment $k$ and select $\omega_b \in N(\omega) - T$ the "best" available move.
    3. Let $\omega \leftarrow \omega_b$. If $f(\omega) < f(\omega^*)$, let $\omega^* \leftarrow \omega$.
    4. If $\omega^*$ is equal to the desired minimum or if $N(\omega) - T = \emptyset$ from 2, stop. Otherwise update $T$ and go to 2.

We did not define the "best" available move at step 2. The simplest —nevertheless powerful— way is to select $\omega_b$ such that $\forall \omega_n \in N(\omega) - T, f(\omega_b) < f(\omega_n)$. This means that the algorithm can select a non-improving move since $f(\omega_b)$ can be greater than $f(\omega^*)$.

The definition of the tabu list (step 4) is also a central one. This list aims at escaping local minima and avoiding cycles. It should then consider as tabu any return to a previous *solution state*. If $s^{-1}$ is the reverse move of $s$, the tabu list can be defined such that $T = \{s_h^{-1} : h > k - t\}$, where $k$ is the iteration index and $t$ defines the size of the time window. Practically, this method is hard to implement, especially because of memory requirement. One usually stores only partial ranges of the moves attributes, which can be shared by other moves. The tabu list then contains collections $C_h$ of moves sharing common attributes: $T = \cup C_h; h > k - t$, where $s_h^{-1} \in C_h$ (Glover, 1989).

Since the tabu list manages moves and not solutions, unvisited solutions can have the tabu status. In order to add flexibility to the research process, *Tabu Search* uses *aspiration levels*. In its simplest form, the aspiration level will allow tabu moves whose evaluation has been the best so far.

Two extra features are usually added to *Tabu Search* (Glover, 1990): *intensification* and *diversification*. These terms can be added to the objective function $\tilde{f} = f + intensification +$

*diversification*.

Intensification aims at closely examining "interesting" areas. The intensification function will favor solutions close to the current best. The simplest way is to get back to a close-to-the-best solution and to reduce the size of the tabu list for some iterations. More sophisticated methods use *long-term memory* memorizing the good components of good solutions.

Diversification aims at avoiding a too local search. The diversification function gives more weight to solutions far from the current one. The simplest way to implement it, is to perform random restarts. One can also penalize the most frequent solutions components.

Let us examine a simple example to illustrate *Tabu Search*.

<<INSERT FIGURE 3>>

The cube (see figure 3) shows the cost and the neighborhood of an eight configurations problem. The random initial configuration is e.g. 10. We will simply define the tabu movements as the reverse movements in each of the three directions, that is if we move along $x+$, the movement $x-$ will be tabu.

First iteration:
- Neighborhood of 10 is 15, 8 and 12.
- The best move is $z+$ which selects 8.
- $z-$ is added to the tabu list.

Second iteration:
- Neighborhood is 11, 13 and 10.
- The best move is $z-$ but it is tabu. The second best move is $x-$ which selects 11.
- $x+$ is added to the tabu list.

Third iteration:
- Neighborhood is 9, 8 and 15.
- The best move is $x-$ but it is tabu. The second best move is $y+$ which selects 9.
- $y-$ is added to the tabu list.

Fourth iteration:
- Neighborhood is 11, 13 and 5.
- The best move is $z-$ which is tabu, but its evaluation is 5 which is lower than the best evaluation so far (8). The aspiration criterion overrides the tabu restriction and the 5 is selected.
- 5 is the global minimum, the research is over.

Despite its simplicity, *Tabu Search* is a highly efficient algorithm. It is known to be one of the most effective meta-heuristics for solving the job-shop scheduling problem (Taillard, 1994; Watson, Whitley, & Howe, 2003). It is used in many different fields like resources planning, financial analysis, logistics, flexible manufacturing…

## *Neural networks*

A neural network is a set of processing units linked by "learnable connections." They are well known in the field of artificial intelligence where they notably provide powerful generalization and clustering tools. Some recurrent neural networks are also useful for optimization. The optimization process is usually based on the minimization of an energy function defined as: $E(x) = E_c(x) + \sum_k a_k E_k(x)$ where $E_c$ is the cost function, $E_k(x)$ are the penalties associated to constraint violations and $a_k$ are the associated weighting parameters. For many optimization problems, the cost function is expressed in a quadratic form $E(x) = -1/2\sum_{i,j} T_{ij} s_i s_j - \sum_i I_i s_i$, where $s_i$ is the signal of the neuron $i$, $T_{ij} = \partial^2 E/\partial s_i \partial s_j$ and $I_i = \partial E/\partial s_i$ (Dreyfus et al., 2002).

Hopfield networks (Hopfield, 1982) are the most famous neural networks used for optimization. They are asynchronous (one randomly selected neuron is updated at each step) fully connected —except self-connection— neural networks (see Figure 4).

<<INSERT FIGURE 4>>

The binary version uses a sign function; the output signal of a neuron is computed as: $s_i = 1$ if $\sum_j w_{ji} s_j - \theta_i \geq 0$, $s_i = 0$ otherwise; where $w_{ji}$ is the weight of the connection between neurons $j$ and $i$, $s_j$ is the signal of the neuron $j$ and $\theta_i$ is the bias (a constant, usually negative, signal).

Such a network is a dynamic system whose attractors are defined by the minima of the energy function defined as:

$$E = -1/2 \sum_{i,j} w_{ij} s_i s_j - \sum_j \theta_j s_j$$

Originally, Hopfield designed his networks as associative memories. Data are stored in the attractors where the network converges starting from partial or noisy data providing a *content-addressable memory*. In 1985, Hopfield demonstrated the optimizing capabilities of his network applying it to the TSP problem (Hopfield & Tank, 1985).

While using Hopfield networks for optimization, the main difficulty is the representation of the problem and the definition of the objective function as the energy of the network. We can illustrate that with the TSP. For $n$ cities, Hopfield and Tank used $n^2$ neurons. A set of $n$ neurons was assigned to each city and the rank of the firing neuron designed the rank of the city during the travel.

<<INSERT TABLE 2>>

Table 2 represents the tour *C-A-E-B-D*. The energy function depends on constraints and cost. For the TSP, the constraints define the validity of the tour that is the fact that each city is visited once. Hopfield and Tank defined the corresponding function as:

$$A/2 \sum_x \sum_i \sum_{j \neq i} V_{xi} V_{xj} + B/2 \sum_i \sum_x \sum_{x \neq y} V_{xi} V_{yi} + C/2 (\sum_x \sum_i V_{xi} - n)^2$$

where $V_{xi}$ is the binary signal of the neuron representing the city $x$ at the position $i$ ($V_i = 0$ or $V_i = 1$) and *A*, *B*, *C* are constant.. The first term is zero when each row contains one "1" (cities are visited once), the second is zero when each column contains one "1" (there is one city per position) and the third term is zero when the matrix contains exactly $n$ "1."

The cost function depends on the length of the tour. It is defined as:

$$D/2 \sum_x \sum_{y \neq x} \sum_i d_{xy} V_{xi} (V_{y,i+1} + V_{y,i-1})$$

where $d_{xy}$ is distance between cities $x$ and $y$ and where $V_{x,n+j} = V_{x,j}$. The energy function is the sum of the four terms. If the constants are large enough ($A = B = 500$, $C = 200$, $D = 500$ in the initial tests), low energy states will correspond to valid tours. The matrix of the connection weights becomes:

$$w_{xi,yj} = -A\delta_{xy}(1-\delta_{ij}) - B\delta_{ij}(1-\delta_{xy}) - C - Dd_{xy}(\delta_{j,i+1} + \delta_{j,i-1})$$

where $w_{xi,yj}$ is the weight of the connection between the neurons representing the city $x$ at position $i$ and the city $y$ at position $j$ and $\delta_{ij} = 1$ if $i = j$, $\delta_{ij} = 0$ otherwise.

The original model of Hopfield-Tank has been quite controversial since their results have proved to be very difficult to reproduce. Thanks to posterior improvements (e.g. *Boltzmann machine* which tries to overcome local minima by using a stochastic activation function (Ackley, Hinton, & Sejnowski, 1985)), Hopfield-Tank model demonstrated its usefulness. It is notably used to solve general (e.g. Gong, Gen, Yamazaki, & Xu, 1995) and quadratic (e.g. Smith, Krishnamoorthy, & Palaniswami, 1996) assignment problems; Cutting stock problems (e.g. Dai, Cha, Guo, & Wang, 1994) or Job-Shop scheduling (e.g. Foo & Takefuji, 1988).

Apart from Hopfield networks, T. Kohonen Self Organizing Maps (SOM) (Kohonen,1997), initially designed to solve clustering problems, are also used for optimization (Smith, 1999), notably since the presentation of the *Elastic Net Method* (Durbin & Willshaw, 1987).They are especially used to solve quadratic assignment problems (Smith, 1995) and vehicle routing problems (e.g. Ramanujam & Sadayappan, 1995).

## *Evolutionary algorithms and Genetic Programming*

Evolutionary algorithms and Genetic Programming have chapters devoted to them in this book, so that this section will remain small and general. Evolutionary algorithms provide a way to solve the following interesting question. Given:
1. a very difficult problem for which no way of finding a good solution is known and where a solution is represented as a set of parameters,
2. a number of previous trials that have all been evaluated.

How can one use the accumulated knowledge to choose a new set of parameters to try out (and therefore do better than a random search)? One could store all the trials in a database and perform statistics on the different parameters characterizing the trials, to try to deduce some traits that will lead to better results. However, in real life, parameters are often interdependent (*epistasis*), so drawing conclusions may not be that easy, even on a large amount of data.

Evolutionary algorithms (EAs) rely on artificial Darwinism to do just that: exploit each and every trial to try out new potential solutions that will hopefully be better than the previous ones: given an initial set of evaluated potential solutions (called a population of individuals), "parents" are *selected* to "give birth" to "children" thanks to "genetic" operators, such as "crossover" and "mutation." "Children" are then evaluated and form the pool of "parents" and "children," a *replacement* operator selects those that will make it to the new "generation."

As can be seen in the previous paragraph, the biological inspiration for this paradigm led to borrow vocabulary specific to this field.

The *selection* and *replacement* operators are the driving force behind artificial evolution. They are biased towards good individuals, meaning that (all in all) the population is getting better along as the generations evolve. A too strong selection pressure will lead to a premature convergence (the population of individuals will converge towards a local optimum) while a too weak selection pressure will prevent any convergence.

Evolutionary algorithms can be used to optimize virtually any kind of problems, even some that cannot be formalized. This makes them usable for interactive problems where the fitness of an individual is given by a human operator (see for example Ian Parmee chapter in this book). They are also very efficient on multi-objective problems, thanks to the fact that they evolve a whole population of individuals at once (see Carlos Coello Coello chapter in this book). Proper techniques (such as NSGA-II (Deb, Agrawal, Pratab, & Meyarivan, 2000)) can be used to create a full Pareto-front in only one run (something impossible to do with *simulated annealing* or *Tabu Search*, for instance).

If EAs can be used for virtually anything, why not try to evolve programs? This is what Genetic Programming (detailed in chapter *Genetic Programming*) is about. Individuals are not merely a set of parameters that need to be optimized, but full programs that are run for evaluation. The main difference between Genetic Programming and standard EAs is that individuals are executed to be evaluated (rather than used in an evaluation function).

*Data-level parallelism*

The powerful data-level parallelism trend appeared in the mid-1980's, with many seminal works (Wolfram, 1984; Rumelhart & McClelland, 1986; Minsky, 1986; Hillis & Steele, 1986; Thinking Machines Corporation, 1986). Object-oriented programming allows to embed intelligent behavior at the data level. The idea is then to put together many small intelligent entities, possibly on parallel machines or even better, a connection machine (cf. CM-1 by D. Hillis of *Thinking Machines*, with 64K processors, 1986).

One of the first applications of the Connexion Machine was to implement particles simulating perfect ball bearings that could move at a single speed in one of six directions, and only connected to their nearest neighbors. The flow of these particles on a large enough scale was very similar to the flow of natural fluids. Thanks to data-level parallelism, the behavior of a complex turbulent fluid —that would have used hours of computation to simulate using Navier-Stokes equations— could be much simply obtained thanks to the parallel evolution of the elementary particles.

The notion of turbulence was not specifically implemented in the behavior of the elementary particles: it just *emerged* when they were put together, and set in motion.

Still in 1986, Craig Reynolds implemented a computer model to simulate coordinated animal motion such as bird flocks and fish schools. He called the creatures *Boids* (Reynolds, July 1987). What he discovered was that one could obtain a flocking or schooling *emergent* behavior by implementing the very simple three following rules into individual boids:
1. Separation: steer to avoid getting too close from local flockmates.
2. Aligment: steer towards the average heading of local flockmates.
3. Cohesion: steer to move towards the average position (center of gravity) of local flockmates.

Note that there is no global knowledge of the position or shape of the flock of boids. Boids only know about the position of their neighbors (local flockmates).

*Particle Swarm Optimization*

In 1975, E. O. Wilson (a sociobiologist) wrote, in reference to fish schooling: "In theory at least, individual members of the school can profit from the discoveries and previous experience of all other members of the school during the search for food" (Wilson, 1975). In 1995, J. Kennedy and R. Eberhart, who were aware of the work of Craig Reynolds saw in this statement that swarms (or fish schools) could possibly be used to find specific places (such as optima) in a search space.

Their first tests on known problems were quite positive, so they refined the model. They quickly saw that rules number 1 and 2 were not needed, so they used Occam's razor to remove them, but visually, this changed the flock behavior into a swarm behavior, hence the *Particle Swarm Optimization* name (PSO) (Kennedy & Eberhart, 1995).

Particles collaborate as a population to find the best possible solution to a problem. Each particle (a potential solution to the problem made of an $n$-dimensions array of parameters) knows of the position of the best solution ever found by the swarm (called *gbest*) and of the best solution it ever found (called *pbest*). However, going directly to either of *pbest* or *gbest* is pointless because these points have already been visited. The idea behind PSO is to have the particles go towards both *pbest* and *gbest* with inertia: it is their speed that is modified, rather than directly their position.

In the original algorithm, each particle `p` has access to:
- its current position in each of the dimensions $i$ of the problem at hand: `p.pos[i]`.
- the best solution it has personally found, i.e. `p.pBestVal` (value of the best found solution) and for each dimension $i$: `p.pBestPos[i]`,
- a velocity for each dimension $i$: `p.Veloc[i]`.
- the best solution found by the particle swarm : `gBest.Val` and for each dimension $i$: `gBest.pos[i]`.

Then, the algorithm runs as follows:
1. Initialise randomly all particles of the population, evaluate them and set their `pBestVal` field to 0.
2. Clone the best particle and copy it into `gBest` and for all particles, set their `pBest` positions to their current positions if their current value is greater than their `pBestVal` value.
3. Calculate new velocities for all particles and for all dimensions:
   ```
   p.Veloc[i]=p.Veloc[i]
   +pIncrement*rand()*(p.pBest[i]-p.Pos[i])
   +gIncrement*rand()*(gBest.pos[i]-p.Pos[i]
   ```
   where `rand()` returns a random value between 0 and 1.
4. Move all particles using the calculated velocities, i.e. for all dimensions of all particles: `p.pos[i]=p.pos[i]+p.Veloc[i]`
5. Evaluate each particle and go back to step 2 until a stopping criterion is met.

In this algorithm, particles are attracted by two locations: the place where they found their personal best result `pBest` and some kind of public knowledge of where the best spot found by the swarm lies `gBest`. Two increments are associated to adjust the velocity towards these particular locations. In their original paper, Kennedy and Eberhardt called `pIncrement` pressure for "simple nostalgia," while `gIncrement` represents pressure towards group knowledge.

Simulations showed that a high value of `pIncrement` relative to `gIncrement` results in excessive wandering of isolated individuals through the problem space, while the reverse (high `gIncrement` *vs* `pIncrement`) results in premature convergence towards a local optimum. Approximately equal values seems to give the best result. A value of 2 was given by Kennedy and Eberhardt to give `rand()` a mean of 1, meaning that agents would have equal chances of moving too fast or too slow towards their aim.

Particle Swarm Optimization uses an elegant and simple algorithm to implement a swarm of individuals evolving in a fitness landscape thanks to which an optimal solution emerges. To quote Kennedy and Eberhardt, "much of the success of particle swarms seems to lie in the agents' tendency to hurtle past their target." The model was later on generalized in (Shi & Eberhart, 1998).

*Ant Colony Optimization*

Ant Colony Optimization is another Nature-inspired algorithm, based on data level parallelism and the concept of emergence. The idea comes from a biology paper (Deneubourg, Pasteels, & Verhaeghe, 1983) which describes the very simple mechanism that ants use to establish an optimal path between a food source and their ant-hill, without central supervision.

This paper came out in 1983, while the data-level parallelism trend was blooming. In 1988, Manderick and Moyson saw the emergent process lying in this description and wrote a seminal paper (Moyson & Manderick, 1988) in which they describe the implementation of virtual ants on a computer. The appendix of this paper contains the equations that govern the behavior of ants, allowing to compute the critical mass of ants above which self organization emerges at the macroscopic level.

In 1991, Colorni et. al describe a *Distributed Optimization by Ant Colonies* (Colorni, Dorigo, & Maniezzo, 1991), and Marco Dorigo's PhD thesis (Dorigo, 1992) uses virtual ants to solve the Traveling Salesman Problem (TSP). Many other papers will follow, describing Ant Colony Optimization (Stützle & Dorigo, 2002; Maniezzo, Gambardella, & Luigi, 2004; Dorigo & Caro, 1999).

Ant colony optimization is based on stigmergy, evaporation and errors. Real or artificial ants communicate by leaving global information in their environment (*stigmergy*) under the form of *pheromones* (odors) that evaporate with time:

- Foraging ants leave the ant-hill with the aim of bringing back food for the community. If there are no pheromones around, they walk randomly. Otherwise, they tend to follow pheromone trails proportionally to the amount of pheromones on the trail.
- If ants find a food source, they detach and carry a bit of the food, and continue walking either randomly, or following a pheromone trail. In either case, ants carrying food leave behind them a trail of pheromones. If they do not follow a pheromone trail and walk randomly, they create a new pheromone trail. If they follow a pheromone trail, they will reinforce it.

The described algorithm is really simple. Robustness and adaptability reside in the volatility of stigmergic information (evaporating pheromones) and the stochastic following of existing paths (error in trail following).

If the food source depletes, ants will not be able to carry food anymore, and will not reinforce the trail leading to the food source. The trail leading to a depleted food source will therefore fade away and disappear automatically, thanks to pheromone evaporation.

If an existing trail is interrupted because of an external cause, ants coming from the food source will start walking randomly around the obstacle, until by chance, the trail leading to the ant-hill is found again. If two alternative solutions are found, the traffic on the shortest one will be greater, meaning that the pheromone scent will be stronger. The shorter trail will therefore appear more attractive to ants coming to the point where they must choose which way they want to go. After a while, the longest path disappears in favor of the shortest.

Solving the TSP with Ant Colony Optimization

An implementation of a TSP solver using ant colony optimization is well described in (Stützle & Dorigo, 1999): Cities are placed on a graph, with edges bearing the distance between the two connected cities. A number of artificial ants are placed on random cities, and move to other cities until they complete a tour. A fitness is computed for each edge of the graph, which is a weighted sum between the amount of pheromones already borne by the edge and the inverse of the distance to the town pointed to by the edge.

At each time step, artificial ants probabilistically choose to follow an edge depending on its fitness value, and deposit an amount of pheromone, to indicate that the edge has been chosen (local updating). This goes on until all ants have completed a tour. When this is done, the ant that has found the shortest tour deposits pheromones on the edges of its tour proportionally to the inverse of the total distance (global updating), and all ants are restarted again from random towns.

This algorithm is very efficient: *HAS-SOP* (Hybridized Ant System for the Sequential Ordering Problem) is one of the fastest known methods to solve Sequential Ordering

Problems (a kind of asymmetric TSP with multiple precedence constraints) (Gambardella & Dorigo, 2000).

# Conclusion

Stochastic optimization has known several eras, where different algorithms techniques have blossomed as they were discovered and used. *Simulated annealing* and neural networks were certainly among the first. They are therefore very well described and widely used in industry and applied sciences.

However, other algorithms (sometimes as ancient as the previously quoted ones) have come of age, thanks to the computation power of modern computers, and should not be ignored, since they bring many advantages over the old ones. Evolutionary algorithms, for instance, are very good at tackling multi-objective problems, since they can provide a complete Pareto-front in only one run. Genetic Programming can be used for symbolic regression problems or to design analog circuits that are human competitive. More recent algorithms, based on the concept of data-level parallelism, allow to cross levels of abstraction in order to find emergent solutions, possibly leading to "a new kind of science" (Wolfram, 2002).

One important thing to keep in mind though is the *No Free Lunch* theorem, which states that no black-box will ever be able to solve any kind of problem better than a random search. The conclusion is that for a specific class of problem, some algorithms will work better than others, and that all these heuristics need to be tailored for a particular application if one really wants to obtain outstanding results.

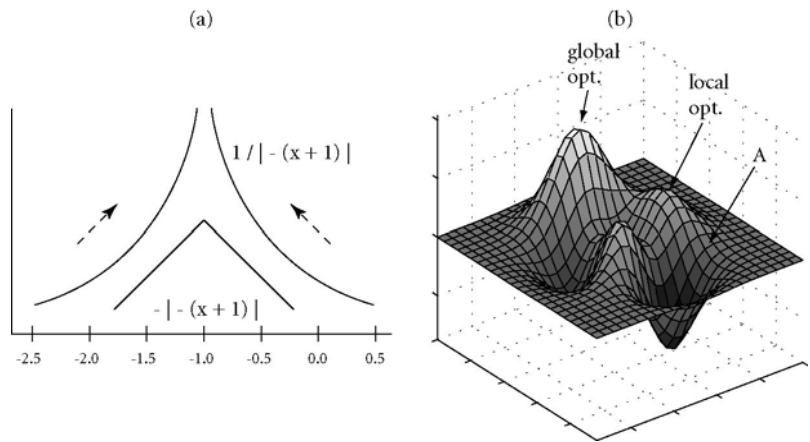

**Figure 1. (a) Fitness landscapes for $x+1=0$. (b) A multimodal fitness landscape.**

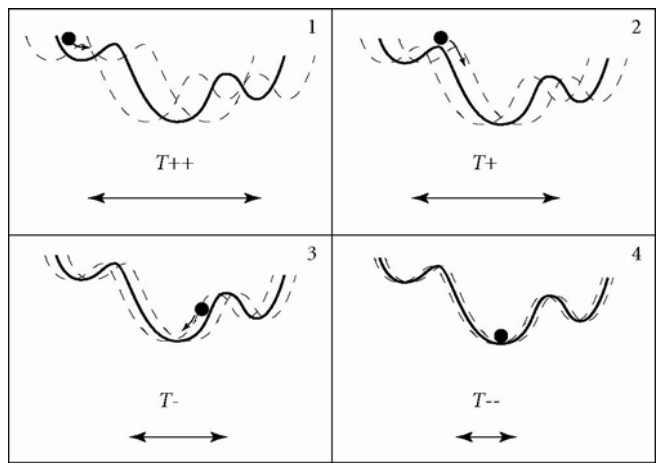

**Figure 2. Simulated Annealing**

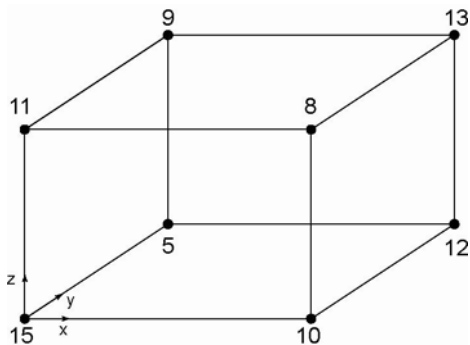

**Figure 3. An illustration of Tabu Search**

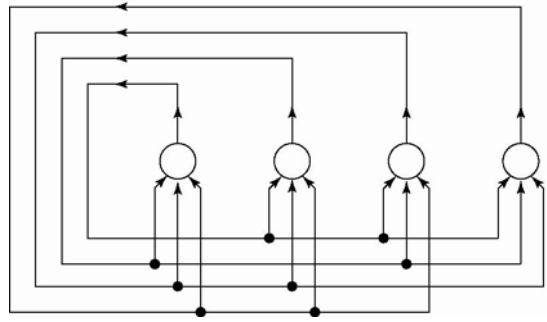

**Figure 4. Hopfield network**

**Table 1: Polynomial vs non-polynomial functions complexity growth[1]**

| O(N) | N=17 | N=18 | N=19 | N=20 |
|---|---|---|---|---|
| $N$ | $17 \times 10^{-9}$ s | $18 \times 10^{-9}$ s | $19 \times 10^{-9}$ s | $20 \times 10^{-9}$ s |
| $N^2$ | $289 \times 10^{-9}$ s | $324 \times 10^{-9}$ s | $361 \times 10^{-9}$ s | $400 \times 10^{-9}$ s |
| $N^5$ | $1.4 \times 10^{-3}$ s | $1.8 \times 10^{-3}$ s | $2.4 \times 10^{-3}$ s | $3.2 \times 10^{-3}$ s |
| $2^N$ | $131 \times 10^{-6}$ s | $262 \times 10^{-6}$ s | $524 \times 10^{-6}$ s | $1 \times 10^{-3}$ s |
| $5^N$ | 12.7 mn | 1 h | 5.29 h | 26.4h |
| TSP | 2.9 h | 2 days | 37 days | 2 years ! |
| $N!$ | 4 days | 74 days | 4 years | 77 years ! |

[1] Considering $10^9$ operations per second, evolution of the algorithm time according to its complexity. TSP stands for *Traveling Salesman Problem*, with complexity $\frac{(N-1)!}{2}$ for $N$ towns (see below).

**Table 2: Hopfield matrix for the TSP**

|   | 1 | 2 | 3 | 4 | 5 |
|---|---|---|---|---|---|
| A | 0 | 1 | 0 | 0 | 0 |
| B | 0 | 0 | 0 | 1 | 0 |
| C | 1 | 0 | 0 | 0 | 0 |
| D | 0 | 0 | 0 | 0 | 1 |
| E | 0 | 0 | 1 | 0 | 0 |